%% file: main.tex
\documentclass[10pt,twocolumn,letterpaper]{article}

\usepackage[pagenumbers]{cvpr} % To force page numbers, e.g. for an arXiv version
\usepackage{multirow}
\usepackage{colortbl}
\usepackage{makecell}
\definecolor{cvprblue}{rgb}{0.21,0.49,0.74}
\usepackage[pagebackref,breaklinks,colorlinks,allcolors=cvprblue]{hyperref}

\def\paperID{618} % *** Enter the Paper ID here
\def\confName{CVPR}
\def\confYear{2025}

\newcommand{\figref}[1]{Fig.~\ref{#1}}
\newcommand{\tabref}[1]{Tab.~\ref{#1}}

\def\eg{\emph{e.g.,~}}
\def\ie{\emph{i.e.,~}}

\newcommand{\nameofmethod}{DFormerv2}

\newcommand{\sArt}{SOTA~}
\newcommand{\myPara}[1]{\vspace{2pt}\noindent\textbf{#1}}

\newcommand{\highlight}[1]{\textbf{\textcolor{BrickRed}{#1}}}

\author{Bo-Wen Yin, ~Jiao-Long Cao, ~Ming-Ming Cheng, ~Qibin Hou\thanks{Qibin Hou is the corresponding author.} \\
VCIP, CS, Nankai University\\
{\tt\small bowenyin@mail.nankai.edu.cn, houqb@nankai.edu.cn}
}

\begin{document}
\title{\nameofmethod{}: Geometry Self-Attention for RGBD Semantic Segmentation}

% \and

\maketitle
% \twocolumn[{%
% \renewcommand\twocolumn[1][]{#1}%
% \maketitle
% \vspace{-15pt}
% \begin{center}
%     \centering
%     \captionsetup{type=figure}   \includegraphics[width=\textwidth]{figs/fig1.pdf}
%     % \put (-362, 53){\footnotesize{{\color{white}{GT}}}}
%     \vspace{-20pt}
%     \captionof{figure}{
%     \small Visualization of our created \textbf{\nameofdataset{}}. Built upon the ImageNet, the \nameofdataset{} contains five popular visual modalities, \ie RGB, Depth, Lidar, Thermal, and Event.}\label{fig:vis_dataset}
% \end{center}%
% }]

%%%%%%%%% ABSTRACT 
\begin{abstract}

Recent advances in scene understanding benefit a lot from depth maps because of the 3D geometry information, especially in complex conditions (\eg low light and overexposed).
Existing approaches encode depth maps along with RGB images and perform feature fusion between them to enable more robust predictions. 
Taking into account that depth can be regarded as a geometry supplement for RGB images, a straightforward question arises: Do we really need to explicitly encode depth information with neural networks as done for RGB images?
Based on this insight, in this paper, we investigate a new way to learn RGBD feature representations and present \nameofmethod{}, a strong RGBD encoder that explicitly uses depth maps as geometry priors rather than encoding depth information with neural networks.
Our goal is to extract the geometry clues from the depth and spatial distances among all the image patch tokens, which will then be used as geometry priors to allocate attention weights in self-attention.
Extensive experiments demonstrate that \nameofmethod{} exhibits exceptional performance in various RGBD semantic segmentation benchmarks.
Code is available at: \href{https://github.com/VCIP-RGBD/DFormer}{https://github.com/VCIP-RGBD/DFormer}.

% first calculate the spatial and depth distance among all the image patches, . 

\end{abstract}

\begin{figure}[tp]
\centering
% \vspace{-0.5cm}
% \setlength{\abovecaptionskip}{-2pt}
\includegraphics[width=0.84\linewidth]{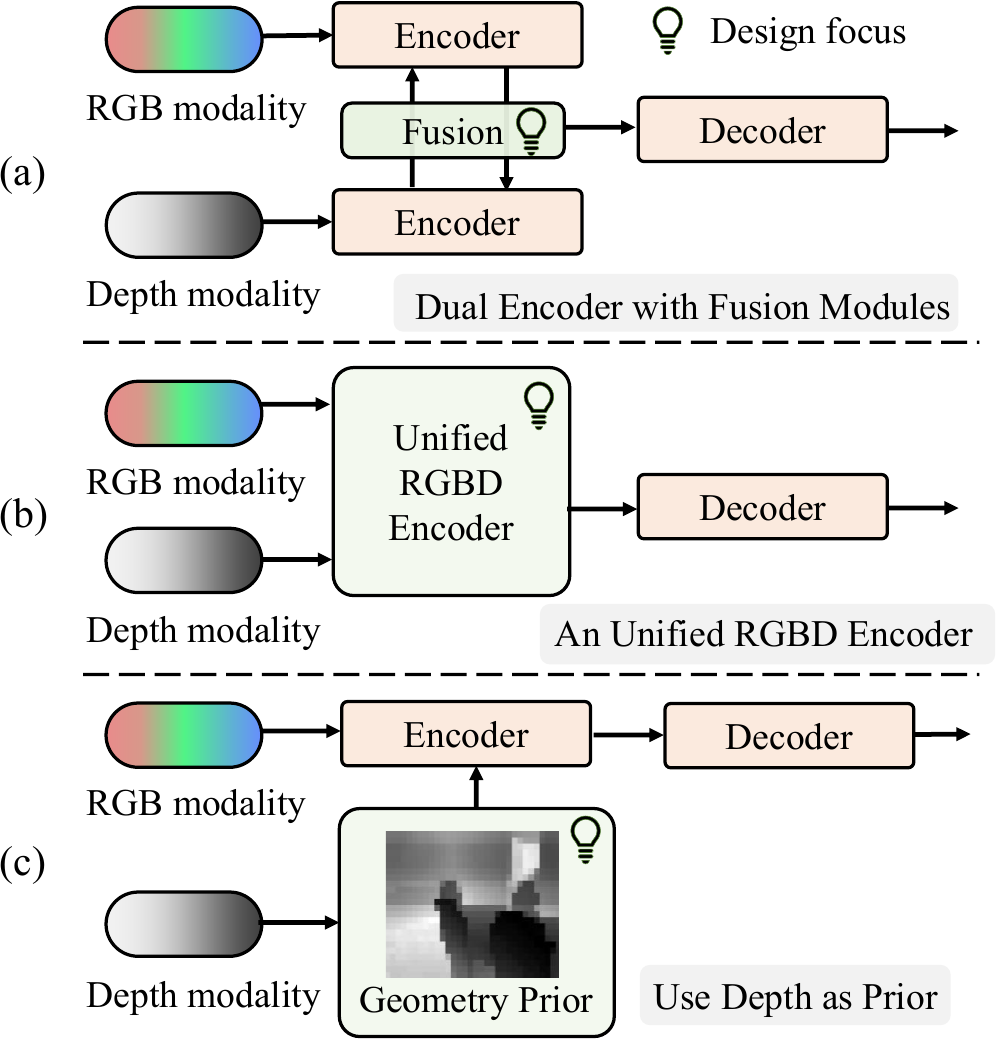}
\vspace{-5pt}
\caption{Comparisons among the main RGBD segmentation pipelines and our approach. (a) Use dual encoders to encode RGB and depth respectively and design fusion modules to fusion them~\cite{zhang2022cmx,jia2024geminifusion}; (b) Adopt an unified RGBD encoder to extract and fuse RGBD features~\cite{bachmann2022multimae,yin2023dformer}; (c) Our \nameofmethod{} use depth to form a geometry prior of the scene and then enhance the visual features.}\label{fig:com_ppl}
\vspace{-2pt}
\end{figure}

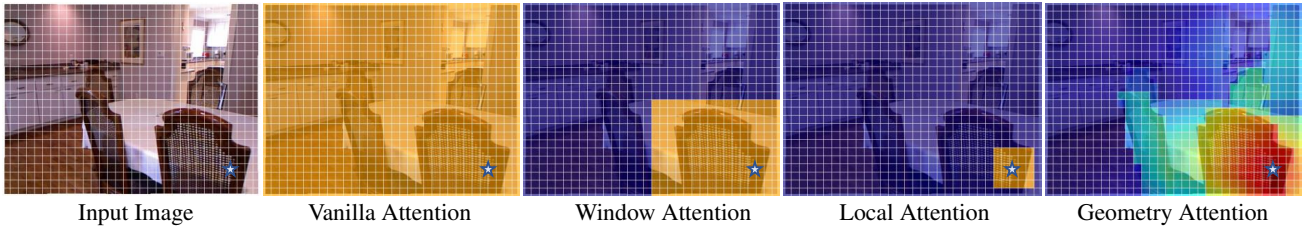
\begin{figure*}[tp]
  \centering
  \includegraphics[width=0.99\linewidth]{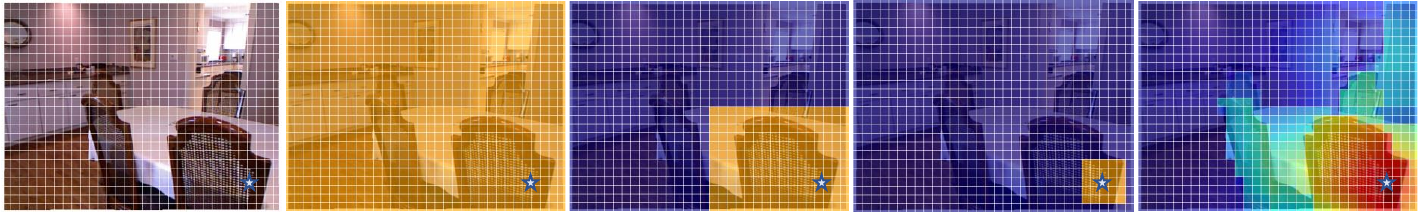}
  \put(-463,-8){\small Input Image}
  \put(-376,-8){\small Vanilla Attention}
  \put(-275,-8){\small Window Attention}
  \put(-175,-8){\small Local Attention}
  \put(-85,-8){\small Geometry Attention}
  \vspace{-5pt}
  \caption{Comparison between geometry self-attention (GSA) and other attention mechanisms, \ie vanilla attention~\cite{dosovitskiy2021vit}, window attention~\cite{liu2021swin,dong2022cswin}, and local attention~\cite{yang2021focal,wang2021crossformer}. The `star' sign means the current query's position. In GSA, colors closer to red represent smaller decay rates, while colors farer away red represent larger ones. In other attention mechanisms, the bright color means the receptive field.}
\vspace{-10pt}
  \label{fig:attn_vis}
\end{figure*}

% \begin{figure}[tp]
% \centering
% % \vspace{-0.5cm}
% % \setlength{\abovecaptionskip}{-2pt}
% \includegraphics[width=0.99\linewidth]{figs/flops_zhexiantu.pdf}
% \vspace{-10pt}
% \caption{Comparisons among the main RGBD segmentation pipelines and our approach.}\label{fig:flops}
% \vspace{-2pt}
% \end{figure}

%%%%%%%%% BODY TEXT
\section{Introduction}

% With the widespread use of modular sensors, multimodal data is becoming increasingly available to access.
Semantic segmentation, aiming at assigning each pixel in an image to a specific pre-defined category label, has been a fundamental area of research in computer vision due to its broad range of applications, such as in intelligent transportation systems and autonomous driving~\cite{jiang2024traffic}.
However, approaches based solely on RGB data often suffer significant performance degradation in complex scenarios, such as cluttered indoor environments or low-light conditions.
In recent years, advancements in 3D modular sensors have made depth data more accessible. 
Integrating RGB-D data makes scene understanding more robust and accurate and thus becomes pivotal in advancing high-level vision tasks.
Furthermore, RGB-D data have demonstrated remarkable potential, surpassing the RGB-based paradigm in various downstream tasks, including autonomous driving~\cite{huang2022multi}, SLAM~\cite{wang2023co}, and robotics~\cite{marchal2020learning}. 

\figref{fig:com_ppl}(a) presents the architecture of current mainstream RGB-D models.
As depicted, it utilizes a dual encoder architecture~\cite{zhang2022cmx, zhang2023delivering}, wherein one encoder extracts features from the RGB modality, while the other processes depth information.
Meanwhile, a fusion strategy is performed to achieve interaction between the information of these two modalities during the encoding process.
Despite the success, the majority existing RGBD segmentation approaches adopt identical backbone architectures to extract features from both RGB and depth data for fusion, neglecting the inherent differences between the RGB and depth.

A series of studies have sought to identify optimal methods for processing depth maps and integrating them with RGB data.
Asyformer~\cite{du2024asymformer} employs a dual-stream asymmetric backbone, \ie using a more efficient encoder for depth data to reduce redundant parameters during feature extraction.
PrimKD~\cite{hao2024primkd} proposes a knowledge distillation (KD)-based method to guide multimodal fusion in RGB-D semantic segmentation, with an emphasis on leveraging the primary RGB modality.
Furthermore, as shown in \figref{fig:com_ppl}(b), DFormer~\cite{yin2023dformer} presents an efficient RGB-D model that encodes both RGB and depth data in a unified encoder via representation learning manners~\cite{he2020momentum,sunprogram}, yet allocates more computational resources to processing the RGB data. 
These methods acknowledge that RGB and depth carry distinct information, each contributing differently to semantic segmentation. 
However, they fail to account for the unique characteristics of the depth modality fully.
In conclusion, how to effectively and efficiently utilize depth information remains an open question and warrants further exploration.

In this paper, taking into account the physical meanings of depth maps that reflect the geometrical information of the given scenes, we consider the way of utilizing depth maps from a new perspective.
Unlike previous works that use neural networks to simultaneously encode RGB images and depth maps as shown in \figref{fig:com_ppl}, we propose directly employing depth maps as geometry priors and using them to guide weight distributions in self-attention, producing a new attention mechanism, called Geometry Self-Attention (GSA).
An illustration of how the proposed GSA works and its differences from other self-attention variants can be found in~\figref{fig:attn_vis}.
In each building block, we model the geometric and spatial relationships among all the patch tokens based on the GSA, a more efficient way to fuse RGB and depth information.
Our method requires no extra layers to process depth maps and hence needs fewer learnable parameters and computations compared to other types of RGBD segmentation methods.
In addition, to reduce the computational burden of vanilla self-attention, we also adopt an axes decomposition operation that decomposes self-attention along both spatial axes of the features. 
% For RGB modality, the appearance of different instances for the same category can be entirely different and the images contain rich data pattern.
% 
% In this way, the geometry the spatial information of pixels can be more efficiently fused, which can help better distinguish semantic categories.
% 
% Thus, we attempt to introduce the depth modality as geometry prior into the self-attention on RGB features to utilize the geometry information within depth, termed as geometry self-attention.
% 
% As shown in \figref{fig:attn_vis}, we compare our geometry self-attention with other Self-Attention mechanisms.

Based on Geometry Self-Attention, we construct a powerful RGB-D vision backbone, called \nameofmethod{}.
We demonstrate the effectiveness of \nameofmethod{} on popular RGB-D semantic segmentation benchmarks, \eg NYU DepthV2~\cite{silberman2012nyu_dataset}, SUNRGBD~\cite{song2015sun_rgbd}, and Deliver~\cite{zhang2023delivering}.
By adding a small decoder head on top of the \nameofmethod{}, our approach sets new state-of-the-art records with less computational cost compared to previous methods.
Remarkably, our base scale model, \nameofmethod{}-B, achieves equal performance with the second-best method Gemnifuision (MiT-B5)~\cite{jia2024geminifusion}, \ie 57.7\% mIoU on NYU DepthV2, with less than half of the computation costs.
Meanwhile, our largest model \nameofmethod{}-L is able to achieve 58.4\% mIoU on NYU DepthV2 with 95.5M parameters. 
Compared with other methods, our \nameofmethod{} achieves the best trade-off between segmentation performance and computations.

Our main contributions can be summarized as follows:

\begin{itemize}
\item To our best knowledge, our work marks the first successful attempt to combine depth information with spatial information as a geometry prior and apply it to the neural network. 

% \item To our best knowledge,  our work marks the first successful attempt to achieve accurate RGB-D segmentation without explicitly encoding depth. 
% We create multimodal ImageNet, \ie \nameofdataset{}, facilitating the multimodal representation learning for many visual modalities. 

\item We propose Geometry Self-Attention which introduces geometry prior to self-attention, to construct an efficient RGB-D encoder, termed \nameofmethod{}.
% We design an effective and flexible interaction method to fuse arbitrary multimodal features.
% 
% It is employed to construct our \nameofmethod{}, which is pretrained on \nameofdataset{} and is appliable to arbitrary RGBX tasks.

\item Our method achieves new state-of-the-art performance with less than half the computational cost of the best current methods on three popular RGB-D semantic segmentation datasets.
% Our \nameofmethod{} sets new \sArt performance across all the RGB-D segmentation benchmarks. 
% \ie RGB + single modality and RGB + multi modalities.

\end{itemize}

\begin{figure*}[tp]
  \centering
\includegraphics[width=0.99\linewidth]{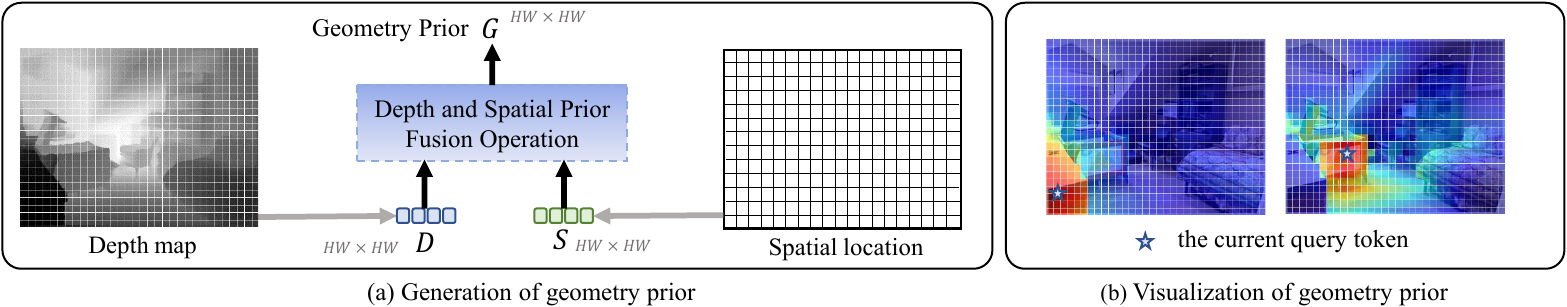}
  \caption{Illustration of the geometry prior. (a) Generation process of the geometry prior. (b) Some visualization of the geometry prior, where the `blue star' means the current query. }
  \vspace{-13pt}
  \label{fig:geo_prior_generation}
\end{figure*}

%---------------------------related work---------------------------------------------

\section{Related Work}

\subsection{RGB-D Semantic Segmentation}

Semantic segmentation~\cite{zhou2017scene}, as one of the core pursuits in computer vision, aims to categorize each pixel in an image into a specific category.
Recently, significant developments in deep learning technologies~\cite{strudel2021segmenter,cheng2021per,cheng2022masked,guo2022segnext,sun2024corrmatch} have been made in this field.
However, some real-world scenes~\cite{xu2021cdada,ji2024segment,gao2022cross,liu2024primitivenet,lin2023sequential} are still challenging to understand using only RGB images, which do not provide sufficient textures, especially in low illumination and fast-moving scenarios.
To address this issue, researchers~\cite{zhou2022canet,zhang2019pattern} propose to utilize depth, which contains 3D geometry information for the scene, to enhance RGB semantic segmentation, known as RGB-D semantic segmentation.
Since then, a series of works have been proposed to achieve the fusion of RGB-D data and leverage the additional information to capture more details.
Here, we delve into the RGB-D fusion schemes and analyze their characteristics.

The current mainstream methods~\cite{hu2019acnet,seichter2021efficient} put a lot of effort into designing interaction modules to fuse the RGB and depth features encoded by two parallel pretrained backbones.
For instance, methods such as CMX~\cite{zhang2022cmx}, TokenFusion~\cite{wang2022multimodal}, GeminiFusion~\cite{jia2024geminifusion} dynamically fuse the RGB-D representations from RGB and depth encoders and aggregate them in the decoder. 
These methods significantly push the performance boundaries in the applications of RGB-D semantic segmentation. 
Nevertheless, they still face two common issues: (1) Treating RGB and depth maps equally with two parallel backbones brings significantly higher computational cost compared to methods based on RGB data; (2) The used backbones are pretrained with RGB images but take an image-depth pair as input during finetuning. The inconsistency between the input causes a huge representation distribution shift.
% 
% To overcome the efficiency issue, the second line of work~\cite{chen2020sa_gate,zhang2023delivering} propose to use lightweight encoding layers without pretrained weight to process the depth.
% % 
% However, due to the random initialized weights for encoding depth and feature fusion, these methods still struggle to leverage the geometry information within depth. 

Recently, DFormer~\cite{yin2023dformer} proposed an RGB-D representation learning framework and utilizes a unified backbone that pretrained on RGB-D pairs to overcome the two issues.
It notices the difference in information density between these two modalities and observes that depth information only requires a small portion of channels to encode.
Although it achieves accurate prediction with high efficiency, it overlooks the intrinsic characteristics of depth modality and merely allocates depth with lower computation cost.
Differently, in this paper, we propose to generate the geometry prior via depth from the perspective of data characteristics.
To the best of our knowledge, this is the first attempt to explicitly use the geometry information of depth without any extra encoding layers.

\subsection{Vision Transformer and Prior Knowledge}
Vision Transformer (ViT)~\cite{dosovitskiy2021vit} was the first to introduce transformer architecture to visual tasks, where images are split into small, non-overlapped patch sequences.
The biggest difference from CNNs~\cite{he2016resnet,hou2022conv2former,liu2022convnet} is that transformers~\cite{liu2021swin,dong2022cswin,cheng2024spt,tang2024wfss} use attention as an alternative to convolution layers to enable global context modeling. 
However, vanilla self-attention incurs a heavy computational burden, as it computes pairwise feature affinities across all patches.
Various sparse attention mechanisms~\cite{han2025agent,zhu2023biformer,xia2022vision,zeng2022not,liu2021swin} have been proposed to alleviate the huge computation cost of self-attention.
At the same time, researchers have presented many studies~\cite{press2021train,sun2024ultra,sun2023retentive,guo2022cmt,wang2022pvt} to incorporate prior knowledge into the transformer model to enhance its representation capacity.
The original transformers~\cite{dosovitskiy2021vit} utilize position encoding to provide positional information for each token.
For vision tasks, swin-transformer~\cite{liu2021swin} proposes to use relative positional encoding instead of the original absolute position encoding.
In contrast, we propose transferring depth into geometry prior knowledge and introducing it into self-attention, termed geometry self-attention.
Compared to the position prior, our geometry prior can model the relationships in the 3D domain across the whole image.

\begin{figure*}[tp]
  \centering
\includegraphics[width=0.99\linewidth]{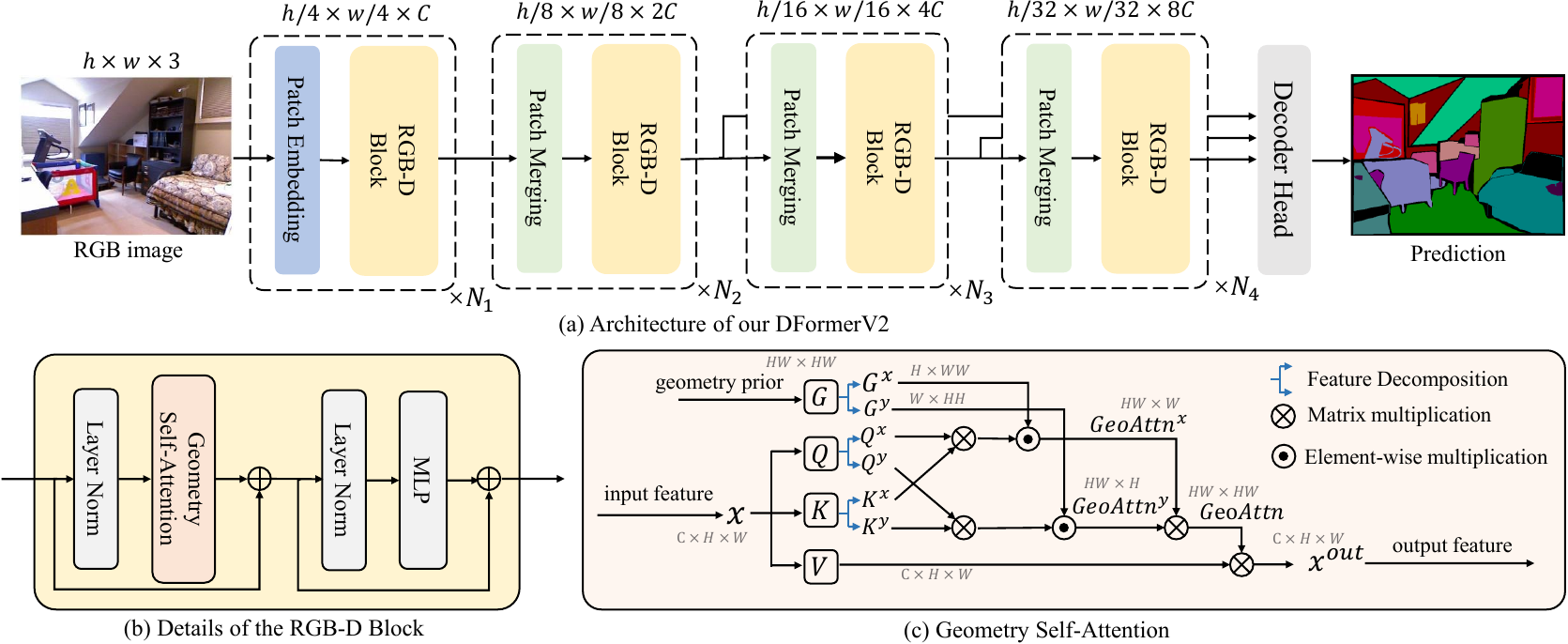}
  \caption{Illustration of our \nameofmethod{}. (a) Overall architecture of our \nameofmethod{}, which contains an encoder with pyramid structure and a decoder head that receives input from the last three stage features. (b) Detailed structure of the basic building block. (c) Detailed illustration of the proposed geometry self-attention mechanism. }
  \vspace{-10pt}
  \label{fig:convert_block}
\end{figure*}

\section{Methodology}

% \subsection{Preliminary}

\subsection{Geometry Prior Generation}

In the vision transformer, the 2D input image of size $h \times w$ is evenly split into $HW$ small patches, where $H$ and $W$ are the numbers of patches per row and column, respectively.
Each patch denoted as $P_{ij}$ is uniquely positioned with a two-dimensional coordinate within the spatial domain, where $i$ and $j$ index the row and column, respectively. 
When the associated depth map is given, the patch in the depth map at the corresponding position reflects its distance from the camera plane.
Based on these two types of priors, we model the geometrical relationships among all the patches and embed them into the self-attention mechanism to form our geometry self-attention.

To be specific, for the depth prior, we perform the average pooling operation for all pixels within the depth patch at position $(i,j)$ to represent its depth location $z_{ij}$ and calculate the distances between each pair of depth patches, which can be defined as:
\begin{equation}
% \begin{align}
    % z_{i} &= \mathrm{AvgPool}(A_{i}), \\
    D_{ij,i'j'} = {|z_{ij}-z_{i'j'}|},
    \label{eq:kl}
% \end{align}
\end{equation}
where $D_{ij,i'j'}$ represents the depth distance between the patches at positions $(i,j)$ and $(i',j')$.
% $i,i'\in \{0,1,...,H-1 \}$, $j,j'\in \left\{0,1,...,W-1 \right\}$, and $HW$ is the number of patches of an image. 
% 
$D$ forms a depth relationship matrix of shape $HW\times HW$.
% 
% \bwy{Considering objects have various poses, different parts of the same object may have significant differences in depth values, \eg the two sides of the table in \figref{fig:attn_vis}.
% % 
% Thus, we take the derivative $D^{\prime}$ of the depth distance $D$ along the location of patches, to better perceive the relationships between object planes.}
% 

The depth relationship matrix $D$ does not contain the spatial distance information, which is also vital to form the geometry clues.
% Given the perception on the object planes, we attempt to integrate it with spatial clues to form geometry priors that model the object relationships of the scene.
Thus, we need to bridge the depth prior with spatial prior as the geometry prior to model comprehensive relationships between image patches.
Similar to the processing of depth prior, we calculate the spatial distance among all the image patches with Manhattan distance.
This can be defined as:
\begin{equation}
\begin{aligned}
    % &z_{i} = AvgPool(P_{i}^{d}), \\
    &S_{ij,i'j'} = {|i-i'|+|j-j'|},
    \label{eq:spatial}
\end{aligned}
\end{equation}
where $S_{ij,i'j'}$ represent the spatial Manhattan distance between patches at positions $(i,j)$ and $(i',j')$. 
Similar to the depth relationship matrix, we can also produce a spatial relationship matrix $S$ of shape ${HW\times HW}$.

Given the depth and spatial distance matrices ${D}$ and ${S}$, we perform the fusion operation to build the bridge between them, as shown in \figref{fig:geo_prior_generation}.
We empirically found that simply using two learnable memories to perform the weighted summation for the depth and spatial priors already works well.
It is worth mentioning that more advanced techniques can also be used to generate the geometry prior by fusing the depth and spatial priors.
We integrate both types of priors to generate the geometry prior ${G}$ of shape ${HW\times HW}$ that stores more comprehensive 3D geometrical relationships for all the image patches.
More visualizations of ${G}$ are shown in \figref{fig:geo_vis_more}.

\subsection{Geometry Self-Attention}
% \myPara{General Form of Self-Attention}
% After splitting an image into $HW$ patches and vision transformers use patch embedding layers to project these patches to features.
% 
Given a feature map $x\in \mathbb{R}^{HW\times C}$,  self-attention can be simply formulated as follows in each head:
\begin{equation}
% \begin{aligned}
     % &Q,K,V = W_{1}x,W_{2}x,W_{3}x, \\
     % &S^{M} = \mathrm{Softmax}(M^{T}S), \\
     % &Q,K,V = W_{1}x,W_{2}x,W_{3}x, \\
     % &S^{M} = \mathrm{Softmax}(M^{T}S), \\
     \mathrm{SelfAtt}(Q,K,V) = \mathrm{Softmax}(QK^{T})V,
    \label{eq:vanilla_attn}
% \end{aligned}
\end{equation}
where $Q,K,V$ are the query, key, and value matrices that can be attained by linear projections.
Inspired by \cite{sun2023retentive,fan2024rmt,shaw2018self} that perform positional encoding to provide spatial information for each token, our geometry self-attention can be achieved by introducing the geometry prior $G$ into the self-attention mechanism via a decay manner.
This process can be written as:
\begin{equation}
% \begin{aligned}
     \mathrm{GeoAttn}(Q,K,V,{G}) = (\mathrm{Softmax}(QK^{T})\odot \beta^{G})V,
    \label{eq:geoattn}
% \end{aligned}
\end{equation}
where $\odot$ means element-wise multiplication, $\beta \in (0,1)$ is the decay rate, and $\beta^{G}$ means taking each element in $G$ as the power of $\beta$ to obtain a new matrix.
% 
% The geometry prior $G$ symbolizes the 3D positional relationships of all the patches and brings the explicit geometry prior into the input feature map.
As $\beta \in (0,1)$ and the elements in $G$ are non-negative numbers, the resulting $\beta^{G} =[ \beta^{g_{ij}} ]_{ij} \in (0,1]^{HW\times HW}$ is a matrix with 1 in diagonal.
Small element values mean long geometric distances.
$\beta^G$ embeds the explicit geometry prior into the attention map via multiplication, and GSA obtains the focus in the near regions, as visualized in \figref{fig:focus_attn}.
Specifically, for a query, the weights of irrelevant key-value pairs are suppressed and the relevant ones are enhanced according to the geometry relationship, benefitting the attention mechanism in modeling intra-object and inter-object relationships.
In practical use, Eqn.~\eqref{eq:geoattn} can also be extended to a multi-head version, and meanwhile, we set different decay rates for different self-attention heads to augment the geometry guidance.

\input{tabs/table_m2}

As demonstrated in previous works for dense prediction tasks~\cite{liu2021swin}, the pyramid structure is often used to encode fine-level features.
However, directly using self-attention to encode high-resolution features
will introduce high computations and memory costs.
Our geometry self-attention also faces this issue.
Thus, inspired by existing sparse attention approaches~\cite{dong2022cswin,hassani2023neighborhood, fan2024rmt,yang2021focal}, we use a simple decomposition manner to perform attention along the horizontal and vertical directions separately, as shown in \figref{fig:convert_block}(c).
To achieve this, we also need to generate the horizontal and vertical geometry priors.
Therefore, we decompose the geometry prior $G$ into $G^{x}$ and $G^{y}$, which reflect the geometry relationship at rows and columns for all the tokens respectively.
Specifically, $G^{y} = [G^{y}_{ij}]_{i=0,1,..,H-1,j=0,1,...,W-1}$ is a matrix of shape $(HW,H)$, where $G^{y}_{ij}$ represents the geometry relationship between the patch at $(i,j)$ and all the patches in $j$-th column.
% 
% The calculation process of the vertical geometry prior is defined as:
% \begin{equation}
%      G^{y}_{j} = \mathrm{Softmax}(D_{j}\cdot M_{j})\cdot \mathrm{Softmax}(M_{j}^{T}\cdot S_{j}),
%     \label{eq:gen_geo_y}
% \end{equation}
% where $G^{y}_{j}$ represents the vertical geometry matrix at the $j$-th column.
% % 
% $G^{y} = [G^{y}_{0},\cdots,G^{y}_{W-1}]$ is a matrix of shape $(H,H,W)$ and reflects the geometry relationship at each column for all the tokens. 
% 
Similarly, we can obtain $G^{x}$ with shape $(HW,W)$.
% , which consisting of $\left \{ G^{x}_{i}\right \}_{i=0,1,2,..,H-1}$.
% 
Then, the calculation of geometry self-attention is formulated as follows:
% \begin{equation}
\begin{align}
     % &Q,K,V = W_{1}x,W_{2}x,W_{3}x, \\
     % &S^{M} = \mathrm{Softmax}(M^{T}S), \\
     % &Q,K,V = W_{1}x,W_{2}x,W_{3}x, \\
     % &S^{M} = \mathrm{Softmax}(M^{T}S), \\
     % &\mathrm{Attention}(Q,K,V) = \mathrm{Softmax}(QK^{T})V, \\
     % D^{w} = \left \{ \gamma{}^{d} \right\},\\
     \mathrm{GeoAttn}^{y} &= (\mathrm{Softmax}(Q^{y}(K^{y})^{T})\odot \beta^{G^{y}}), \\
     \mathrm{GeoAttn}^{x} &= (\mathrm{Softmax}(Q^{x}(K^{x})^{T})\odot \beta^{G^{x}}), \\
     \mathrm{GeoAttn} &= \mathrm{GeoAttn}^{y}(\mathrm{GeoAttn}^{x}V)^{T},
    \label{eq:geo_des_attn}
\end{align}
% \end{equation}
where $Q^{y}(K^{y})^{T}$ and $Q^{x}(K^{x})^{T}$ means perform attention calculation along vertical and horizontal axis.

\subsection{\nameofmethod{} Architecture}
\figref{fig:convert_block} illustrates the overall architecture of \nameofmethod{}, which follows the widely-used encoder-decoder framework.
The encoder is composed of four stages, which are utilized to produce multi-scale features.
Each stage contains a stack of geometry self-attention blocks.
The first three stages perform decomposition on geometry self-attention, while the last one does not.
A lightweight decoder head is employed to transform these visual features into RGB-D semantic segmentation results.
 
Given an RGB image, it is first processed by a stem layer, consisting of two convolutions with kernel size $3 \times 3$ and stride 2.
Then, the RGB features are fed into the hierarchical encoder to encode multi-scale features at $\{ 1/4, 1/8, 1/16, 1/32 \}$ of the original image resolution.
Different from existing methods, there is no need for our \nameofmethod{} to explicitly encode depth maps.
We just need to perform the average pooling operation with different pooling kernels and strides on the depth map to the four scales corresponding to the geometry self-attention blocks in the encoder and then utilize them to generate the geometry prior for each block. 
Based on the configurations of the geometry self-attention blocks in each stage, we design a series of encoder variants, termed  \nameofmethod{}-S, \nameofmethod{}-B, and \nameofmethod{}-L, respectively, with the same architecture but different model sizes. 
Detailed configurations of these variants can be found in the supplementary materials.

% We implement them with transformer attention mechanism and large kernel convolution.

% \subsection{} 

%----------------------------- experiments-------------------------------------------

\section{Experiments}

\begin{figure}[tp]
\centering
% \vspace{-0.5cm}
% \setlength{\abovecaptionskip}{-2pt}
\includegraphics[width=0.99\linewidth]{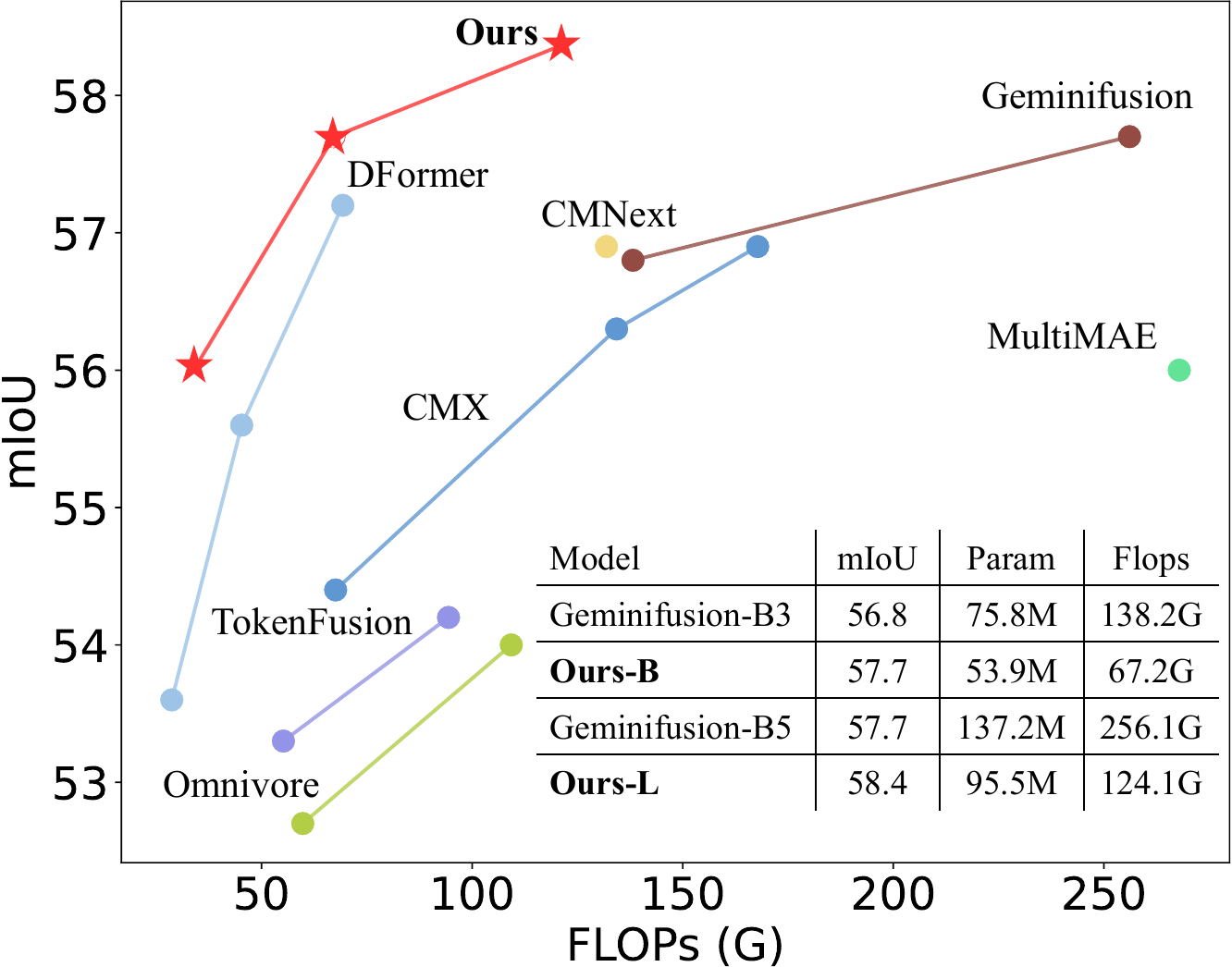}
\vspace{-8pt}
\caption{Performance-computation comparisons between our \nameofmethod{} and other \sArt methods on NYU DepthV2~\cite{silberman2012nyu_dataset}.}\label{fig:zhexian}
\vspace{-5pt}
\end{figure}

\subsection{Implementation Details}

\myPara{Pretraining settings.}
Following DFormer~\cite{yin2023dformer} and MultiMAE~\cite{bachmann2022multimae}, we perform  RGB-D pretraining on ImageNet-1K for our \nameofmethod{}, to endow the encoder with the ability to achieve the interaction between RGB and depth modalities and generate transferable representations with rich semantic and spatial information.
The depth maps for ImageNet are generated by depth estimation method~\cite{yin2023dformer}.
The standard cross-entropy loss is employed as our optimization objective, and the number of training epochs is set to 300, like most pretrained models~\cite{liu2022convnet}.
Following previous works~\cite{yin2023dformer,zhang2022cmx}, the AdamW~\cite{kingma2014adam} with learning rate 1e-3 and weight decay 5e-2 are adopted, and we set the batch size to 1024. 
More detailed settings for each variant of \nameofmethod{} are described in the supplementary materials.

\myPara{Datasets and settings for finetuning.}
Following the commonly used experiment settings of RGB-D semantic segmentation works~\cite{hao2024primkd, jia2024geminifusion, yin2023dformer}, we evaluate our \nameofmethod{} on two popular datasets, \ie NYU DepthV2~\cite{silberman2012nyu_dataset} and SUNRGBD~\cite{song2015sun_rgbd}.
Additionally, we conduct experiments on the Deliver dataset~\cite{zhang2023delivering}, as done in~\cite{jia2024geminifusion}.
In line with DFormer~\cite{yin2023dformer}, we use a lightweight head~\cite{geng2021attention} as our decoder to build our RGB-D semantic segmentation model. 
We only adopt two simple data augmentation methods, \ie random horizontal flipping and random scaling (from 0.5 to 1.75), when finetuning models.
We use the cross-entropy loss as the optimization objective and AdamW~\cite{kingma2014adam} as our optimizer with
an initial learning rate of 6e-5 and the poly decay schedule.
For the NYU DepthV2 and SUNRGBD datasets, we crop and resize the images to $480\times 640$ and $480\times 480$ respectively for training.
During the evaluation, we adopt the mean Intersection over Union (mIoU), which is averaged across all semantic categories, as the primary evaluation metric to measure the segmentation accuracy.
Following recent works~\cite{zhang2022cmx, yin2023dformer, zhang2023delivering}, we employ multi-scale (MS) flip inference strategies at scales $\left\{ 0.5,0.75,1,1.25,1.5\right\}$.
We adopt the same training and testing strategy as CMNeXt~\cite{zhang2023delivering} on DeLiVER, where the images are resized to $1024\times 1024$. 
More details can be found in the supplementary materials.

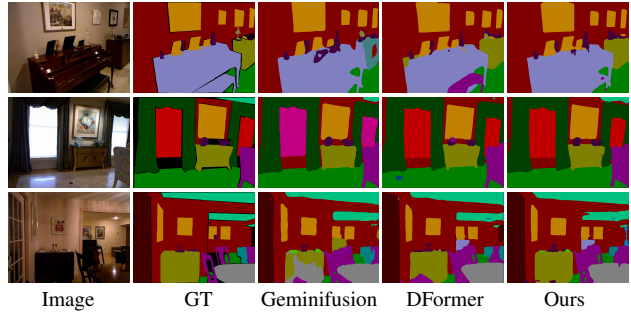
\begin{figure}[tp]
\centering
% \vspace{-0.5cm}
% \setlength{\abovecaptionskip}{-2pt}
\includegraphics[width=0.99\linewidth]{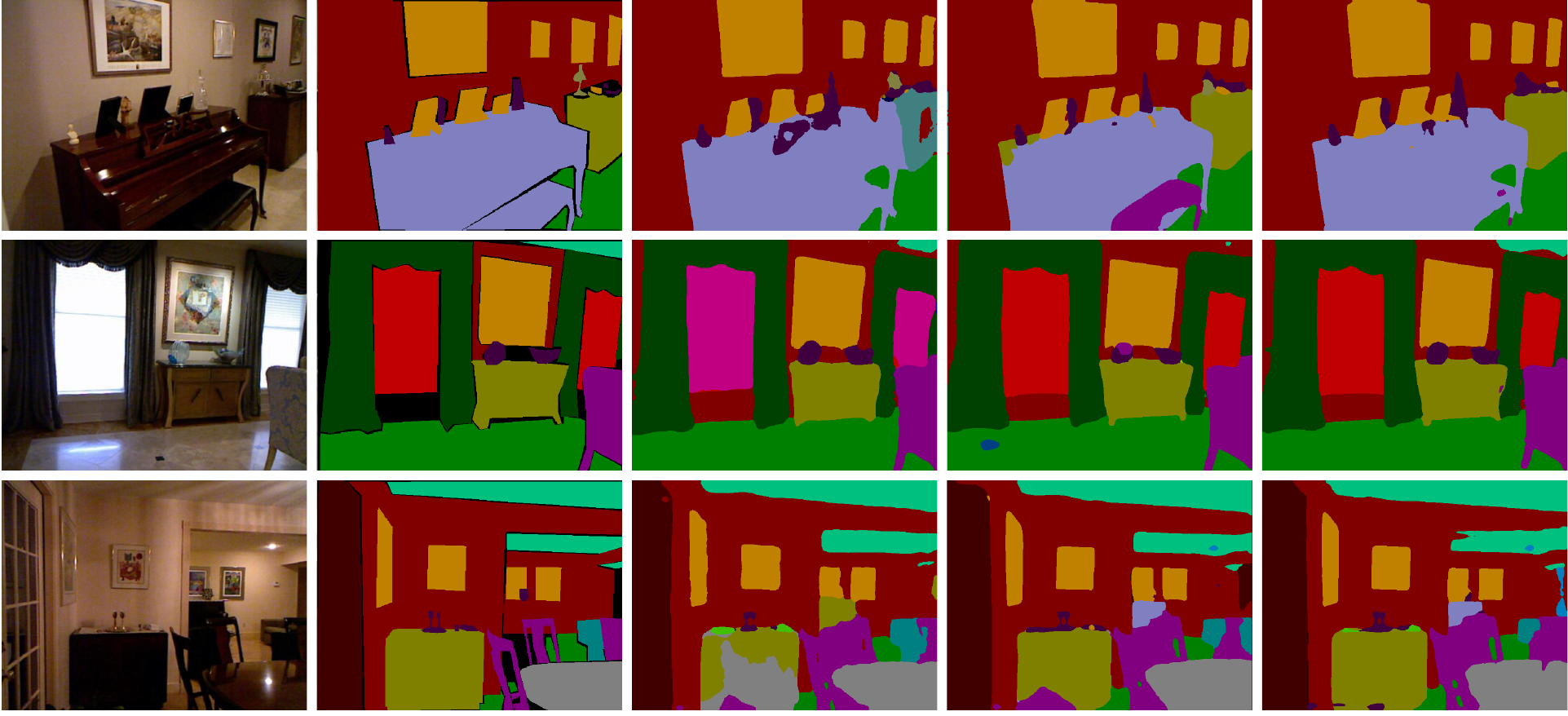}
\put (-222, -8){\footnotesize{{\color{black}{Image}}}}
\put (-168, -8){\footnotesize{{\color{black}{GT}}}}
\put (-139, -8){\footnotesize{{\color{black}{Geminifusion}}}}
\put (-84, -8){\footnotesize{{\color{black}{DFormer}}}}
\put (-32, -8){\footnotesize{{\color{black}{Ours}}}}
\vspace{-5pt}
\caption{Qualitative comparisons with GeminiFusion-B5~\cite{jia2024geminifusion} and DFormer-L~\cite{yin2023dformer}. `GT' is the ground truth.}\label{fig:visualize}
\vspace{-7pt}
\end{figure}

\subsection{Comparisons with Other Methods} \label{subsec:quan}
We compare our \nameofmethod{} with 17 recent RGB-D semantic segmentation approachs on the NYU DepthV2~\cite{silberman2012nyu_dataset}, SUNRGBD~\cite{song2015sun_rgbd}, and Deliver~\cite{zhang2023delivering} datasets.
In \tabref{tab:rgbd_sota}, we categorize the variants of all methods into three sets based on model scale, \ie small scale, base scale, and large scale, for a more intuitive and fair comparison. 
As can be seen, \nameofmethod{} achieves new \sArt performance across all the model scale settings on the two benchmarks.
We also plot the performance computation cost curves of different methods in \figref{fig:zhexian}.
\nameofmethod{} achieves better performance and computation trade-off compared to other methods.
Specifically, our largest model, \ie \nameofmethod{}-L achieves 58.4$\%$ mIoU with 95.5M parameters and 124.1G Flops, surpassing the second-best method Gemnifusion~\cite{jia2024geminifusion} by 0.7\% with less than its half computations.
Similarly, at base and small scales, our \nameofmethod{} also consistently outperforms other \sArt methods with higher efficiency. 
In SUNRGBD and Deliver (\tabref{tab:deliver}) datasets, our \nameofmethod{} also brings significant improvements.
Moreover, the visual comparisons between the semantic segmentation results of our \nameofmethod{} and Gemnifusion~\cite{jia2024geminifusion} are shown in \figref{fig:visualize}.
These improvements demonstrate that our \nameofmethod{} can more efficiently utilize the geometry prior within the depth maps without explicit encoding, and hence yields more accurate predictions with even lower computational cost.

\begin{table}[tp]
  \centering
% \vskip -1ex
    % \setlength\tabcolsep{6pt}
    % \vskip -1ex
    \setlength{\tabcolsep}{5pt}
    \footnotesize
    \vspace{-10pt}
    \centering
    \renewcommand{\arraystretch}{1.0}
    	\begin{tabular}{lccccc}
        \toprule
        Model &Backbone   & Params & \textbf{Flops}& \textbf{mIoU} \\
        \midrule\midrule
        % DFormer$_{24}$~\cite{yin2023dformer}&DFormer-L&39.0M&69.3G&57.2 \\
        HRFuser~\cite{broedermann2022hrfuser}&HRFormer-T&30.5M&223.0G&51.9\\
        TokenFusion~\cite{wang2022multimodal}&MiT-B2&26.0M&55.0G&60.3\\
        \rowcolor{gray!15}\highlight{$\bigstar$} \nameofmethod{}-S&\nameofmethod{}-S&26.7M &28.9G&\highlight{63.7}\\ \midrule
        CMX~\cite{zhang2022cmx}&MiT-B2&66.6M&65.7G&62.7\\
        CMNext~\cite{zhang2023delivering}&MiT-B2&58.7M&62.9G&63.6\\
         \rowcolor{gray!15}\highlight{$\bigstar$} \nameofmethod{}-B&\nameofmethod{}-B&53.9M &60.8G&\highlight{65.2}\\ \midrule
         CMNext~\cite{zhang2023delivering}&MiT-B4&116.6M&112.0G&66.3\\
        GeminiFusion$_{\rm 24}$~\cite{jia2024geminifusion}    & MiT-B5      &  137.2M        &       218.4G             & 66.9                                         \\ 
        TokenFusion~\cite{wang2022multimodal}&MiT-B5&83.3M&144.7G&63.5\\
         \rowcolor{gray!15}\highlight{$\bigstar$} \nameofmethod{}-L&\nameofmethod{}-L&95.5M &114.5G&\highlight{67.1}\\
        \bottomrule
        \end{tabular}
    \hspace{\fill}
    \hspace{\fill}
    \vspace{-8pt}
    \caption{\small Results on Deliver~\cite{zhang2023delivering} dataset. Following~\cite{zhang2023delivering}, the Flops is calculated on the images with shape $512\times 512$.}\label{tab:deliver}
    % \vspace{-10pt}
\end{table}

\begin{figure}[tp]
  \centering
  \includegraphics[width=0.99\linewidth]{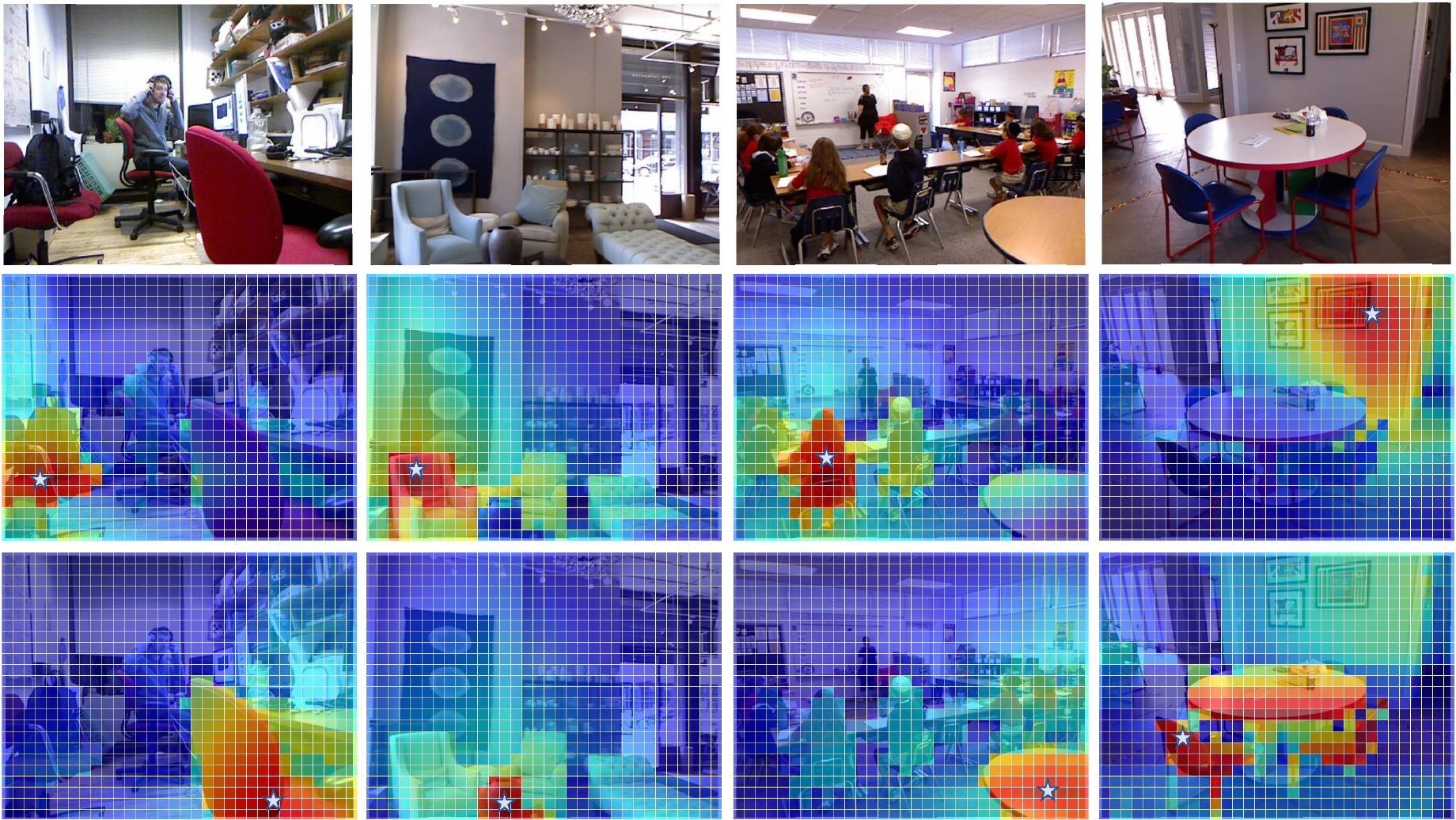}
  \vspace{-5pt}
  \caption{Some visualization samples of the geometry prior. The blue `star' means the current query token. It is important to note that the visualized prior is only obtained from the depth map.}
\vspace{-10pt}
  \label{fig:geo_vis_more}
\end{figure}

\subsection{Model Analysis} \label{subsec:abl}
% \myPara{A roadmap from ViT to \nameofmethod{}.}

\myPara{Geometry self-attention.}
The proposed geometry self-attention consists of depth prior, spatial prior, and priors fusion, which are integrated into a unified geometry prior. 
To evaluate the effectiveness of each component, we present a roadmap from the vanilla self-attention to the geometry self-attention in \tabref{tab:geo_ablation}.
First of all, we introduce the depth prior, spatial prior into the vanilla self-attention (steps 1-2), respectively, to observe the impact on performance. 
These substitutions result in accuracy improvement of 2.6\%, and 1.8\% on NYUDepthV2 and 1.7\%, and 1.3\% on SUNRGBD, respectively, compared to the baseline, highlighting the importance of incorporating these priors in self-attention.
However, it is also apparent that the simple addition of depth and spatial priors results in only a modest improvement over using just the depth prior, indicating that this method of integration may not be effective.
In Step 3, when we introduce fusion operation to bridge the two priors and form the geometry prior, we observe a further improvements on NYUDepthV2 and SUNRGBD, with a negligible increase in computational cost. 
These results (step 1-3) shows that the geometry prior significantly enhances performance with little increase in complexity.
Moreover, in Step 4, we perform a decomposition of the attention mechanism, which further alleviates the computational burden while maintaining almost the same level of performance.
Overall, compared to self-attention, integrating geometry priors enables better RGB-D segmentation with minimal computational overhead and a slight increase in parameters.

\begin{table}[tp]
  \centering
% \vskip -1ex
    % \setlength\tabcolsep{6pt}
    % \vskip -1ex
    \setlength{\tabcolsep}{1.5pt}
    \footnotesize
    \vspace{-10pt}
    \centering
    \renewcommand{\arraystretch}{1.0}
    	\begin{tabular}{ccccccc}
        \toprule
        Step & Attention   & Params & Flops & NYUDepthV2 &SUNRGBD \\
        \midrule\midrule
        0&Vanilla Attn&26.5M&51.4G&51.7&47.8\\ 
        1&+Only Depth Prior&26.5M&51.4G&54.3 ({\highlight{+2.6}})&49.9 ({\highlight{+1.7}})\\ 
        2&+Only Spatial Prior&26.5M&51.4G&53.5 ({\highlight{+1.8}})&49.1 ({\highlight{+1.3}})\\
        3&+Both Priors&26.7M&51.7G&56.2 ({\highlight{+4.5}})&51.7 ({\highlight{+3.9}})\\
        \rowcolor{gray!15}4&+decomposition&26.7M &33.9G&56.0 ({\highlight{+4.3}})&51.5 ({\highlight{+3.7}})\\
        % 5&DFormer$_{24}$~\cite{yin2023dformer}&29.5M&45.3G&55.6\\
        \bottomrule
        \end{tabular}
    \hspace{\fill}
    \hspace{\fill}
    \vspace{-8pt}
    \caption{\small The ablation experiments demonstrate the full roadmap from vanilla Self-Attention to our geometry self-attention on the small scale of \nameofmethod{}. In step 0 and 2, we only input RGB images while we use RGB-D at all the other steps.}\label{tab:geo_ablation}
    % \vspace{-10pt}
\end{table}

\begin{figure}[t]
\vspace{1pt}
\centering
    \setlength{\abovecaptionskip}{2pt}
    \includegraphics[width=0.99\linewidth]{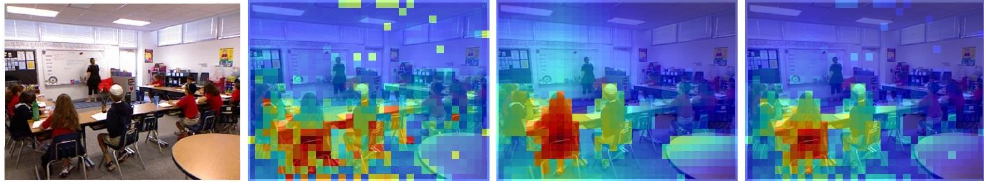}
    \put(-215,-8){\footnotesize Image}
    \put(-170,-8){\footnotesize Attention map}
    \put(-113,-8){\footnotesize Geometry prior}
    \put(-56,-8){\footnotesize Focused attention}
    % \vspace{-10pt}
    \caption{Visualization of the focused attention.}
    \label{fig:focus_attn}
    % \vspace{-10pt}
\end{figure}

\begin{figure}[tp]
  \centering
  \includegraphics[width=0.99\linewidth]{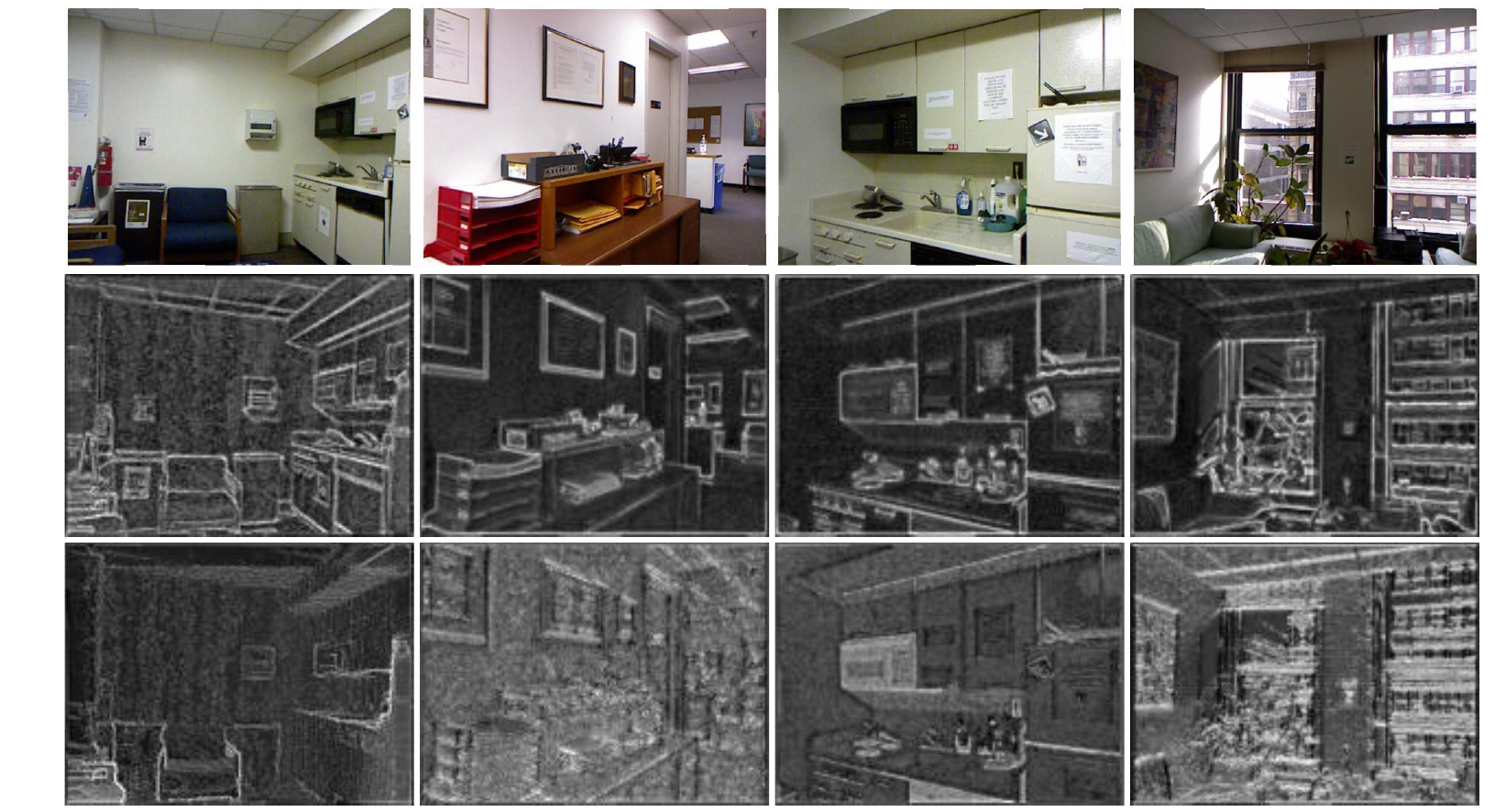}
  \put (-233,100){\rotatebox{90}{\footnotesize{image}}}
  \put (-233,50){\rotatebox{90}{\footnotesize{w/ prior}}}
  \put (-233,8){\rotatebox{90}{\footnotesize{w/o prior}}}
  \vspace{-5pt}
  \caption{Visualization of the features with and without geometry priors. They are randomly picked from the first stage output.}
  \label{fig:fea_map}
  \vspace{-8pt}
\end{figure}

\myPara{More insights about the geometry prior.}
The geometry prior $G$ with a shape $HW\times HW$ is derived from the depth map and represents the geometric relationships between each pair of tokens. 
To provide more insights into this prior, we randomly select several tokens and visualize their geometry relationships with other tokens in \figref{fig:geo_vis_more}. 
For each query token, the geometry prior accurately identifies the object to which it belongs and captures the geometric relationship between this object and its nearby counterparts.
% 
% For example, at the bottom row of the 1st column in \figref{fig:geo_vis_more}, the geometry prior shows that the highlighted token belongs to the chair and that this object is closely related to the nearby table.
% 
The perception of the geometry relationship between objects can help our model better distinguish different semantic objects, for example, chairs are often under the table and people often sit on chairs. 
The focused attention within our GSA is visualized in \figref{fig:focus_attn}.
Introducing geometry prior to the self-attention mechanism enables the model to better understand the geometric structures of objects and the position relationship between objects in complex scenes resulting in more accurate segmentation results.
% 
% \begin{figure}[tp]
%   %\setlength{\belowcaptionskip}{-0.5cm}
%   \centering
%   \includegraphics[width=0.99\linewidth]{figs/attn_vis_more.pdf}
%   \vspace{-5pt}
%   \caption{Some visualization samples of the geometry prior. The blue `star' means the current query token. It is important to note that the visualized prior is only obtained from the depth map.}
% \vspace{-10pt}
%   \label{fig:geo_vis_more}
% \end{figure}
% 
Additionally, we visualize the feature maps with and without the use of geometry priors, as shown in \figref{fig:fea_map}.
It can be seen that introducing geometry priors can help our model better capture the object's details and improve the segmentation performance.

\begin{table}[tp]
  \centering
% \vskip -1ex
    % \setlength\tabcolsep{6pt}
    % \vskip -1ex
    \setlength{\tabcolsep}{7pt}
    \footnotesize
    % \vspace{-10pt}
    \centering
    \renewcommand{\arraystretch}{1.0}
    	\begin{tabular}{lccccc}
        \toprule
         Method   & Params & Flops & NYUDepthV2 &SUNRGBD \\
        \midrule\midrule
        Conv&26.9M&34.3G&55.8&51.3\\ 
        Addition&26.5M&33.5G&54.6&50.4\\
        Hadamard&26.5M&33.6G&54.9&50.9\\
        \rowcolor{gray!15}Memory&26.7M&33.9G&56.2 &51.7\\ 
        % \\
        % \rowcolor{gray!15}+decomposition&26.7M &33.9G&56.0 \\
        % 5&DFormer$_{24}$~\cite{yin2023dformer}&29.5M&45.3G&55.6\\
        \bottomrule
        \end{tabular}
    \hspace{\fill}
    \hspace{\fill}
    \vspace{-5pt}
    \caption{\small Different operations to bridge the depth prior and spatial prior on our small scale model.}\label{tab:mem_ablation}
    % \vspace{-5pt}
\end{table}

\begin{table}[tp]
  \centering
% \vskip -1ex
    % \setlength\tabcolsep{6pt}
    % \vskip -1ex
    \setlength{\tabcolsep}{9pt}
    \footnotesize
    % \vspace{-10pt}
    \centering
    \renewcommand{\arraystretch}{1.0}
    	\begin{tabular}{ccc}
        \toprule
         Settings   & NYUDepthV2 &SUNRGBD \\
        \midrule\midrule
        fixed to 0.25 for all heads&55.7&51.1\\
        fixed to 0.5 for all heads&55.5&51.0\\
        fixed to 0.75 for all heads&55.7&51.2\\
        linearly sampled in [0.5, 1.0)&55.9&51.5\\
        \rowcolor{gray!15}linearly sampled in [0.75, 1.0)&56.0&51.5\\
        \bottomrule
        \end{tabular}
    \hspace{\fill}
    \hspace{\fill}
    \vspace{-8pt}
        \caption{\small Effect of different decay strategies in geometry self-attention on \nameofmethod{}-S.}\label{tab:decay}
    % \vspace{-13pt}
\end{table}

\myPara{Fusion operation.}
To build the bridge between depth prior and spatial prior and form the geometry prior, we leverage memory weights to form the geometry clues from the depth and spatial distances among all the image patch tokens.
To validate the effectiveness of the fusion operation, we also use some other operations to replace it, including addition, Hadamard product,  and convolutional layers.
As shown in \tabref{tab:mem_ablation}, we can see that the memory weights leads to better results than other operations.

% \myPara{About Data Format of the Prior.}

\myPara{Decay Rate.} 
When introducing geometry prior into the attention mechanism as formulated in Eq.~\eqref{eq:geoattn}, we use a decay rate $\beta$ to control the extent of the prior's influence on the features.
Here, we investigate how the model performance would change when different decay rate strategies are adopted. 
The results can be found in \tabref{tab:decay}. 
It indicates that assigning distinct decay rates $\beta$ to different heads in our geometry self-attention introduces multi-scale enhancement and more diversity, which further benefit performance.
Thus, we sample the value of $\beta$ in $[0.75,1.0]$ by default.

% \myPara{Analysis about the efficiency of \nameofmethod{}.}

\begin{table}[t]
    \centering
    \footnotesize
    \setlength{\tabcolsep}{1.3mm}
    \begin{tabular}{lcccc}
        \toprule[1pt]
        Model & \makecell{Params} & \makecell{FLOPs} & \makecell{Latency$\downarrow$} & \makecell{NYU DepthV2}\\
        \midrule[0.5pt]
        Omnivore~\cite{girdhar2022omnivore}&29.1M&32.7G&40.1ms&49.7\\
        DFormer-B~\cite{yin2023dformer}&29.5M&41.9G&42.8ms&55.6\\
        \rowcolor{gray!15}\nameofmethod{}-S & 26.7M & 33.9G & 43.9ms & 56.0 \\
        \midrule[0.5pt]
        % Swin-B & 88 & 15.5 & 756 & 83.5\\
       CMX-B2~\cite{zhang2022cmx} & 66.6M & 65.6G & 71.5ms & 54.4 \\
        DFormer-L~\cite{yin2023dformer} & 39.0M & 69.3G & 44.5ms & 57.2 \\
        GeminiFusion-B3~\cite{jia2024geminifusion} & 75.8M & 138.2G & 68.2ms & 56.8\\
        \rowcolor{gray!15}\nameofmethod{}-B & 53.9M & 67.2G & 50.7ms & 57.7 \\
        \midrule
        CMX-B5~\cite{zhang2022cmx}&181.1M&167.8G&114.9ms&56.9\\
        CMNext-B4~\cite{zhang2023delivering}&119.6M&131.9G&98.5ms&56.9\\
        MultiMAE~\cite{bachmann2022multimae} & 95.2M & 267.9G & 76.9ms & 56.0 \\
        GeminiFusion-B5~\cite{jia2024geminifusion} & 137.2M & 256.1G & 108.7ms & 57.7 \\
        \rowcolor{gray!15}\nameofmethod{}-L & 95.5M & 124.1G & 79.9ms & 58.4\\
        \bottomrule[1pt]
    \end{tabular}
    \vspace{-8pt}
    \caption{\small Comparison of the inference latency between our method and recent SOTA models. `$\downarrow$': the lower the better.} \label{tab:eff1}
    % \vspace{-5mm}
    % \vspace{-8pt}
\end{table}

% \begin{table}[t]
%     \centering
%     \caption{\small Comparison of the inference latency between our \nameofmethod{} and recent SOTA models.}
%     % \vspace{-5mm}
%     \vspace{-5pt}
%     \label{tab:sal1}
% \begin{tabular}{lcccccccc}
%         \toprule
%         \textbf{\multirow{2}{*}{Model}}  &\textbf{\multirow{2}{*}{Backbone}}   & \textbf{\multirow{2}{*}{Params}} &\multicolumn{3}{c}{NYUDepthv2} & \multicolumn{3}{c}{SUN-RGBD} \\ \cmidrule(lr){4-6}\cmidrule(lr){7-9}
%          && &\textbf{Input size}& \textbf{Flops}& \textbf{mIoU}   &\textbf{Input size}& \textbf{Flops}& \textbf{mIoU} \\
%         \midrule\midrule
%         ESANet$_{\rm 21}$~\cite{seichter2021efficient}&ResNet-34&31.2M&$480\times 640$&34.9G&50.3&$480\times 640$&34.9G&48.2\\ 
%         TokenFusion$_{\rm 22}$~\cite{wang2022multimodal}& MiT-B2&26.0M&$480\times640$&55.2G&53.3&$530\times 730$&71.1G&50.3\\ 
% \end{tabular}
% \end{table}

\begin{table}[t]
    \centering
    \footnotesize
    \setlength{\tabcolsep}{3mm}
    \begin{tabular}{lcccc}
        \toprule[1pt]
        \textbf{\multirow{2}{*}{Modality}}  &\textbf{\multirow{2}{*}{Params}}    &\multicolumn{1}{c}{Classification} & \multicolumn{2}{c}{Segmentation} \\ \cmidrule(lr){3-3}\cmidrule(lr){4-5}
         & &Top-1 Acc$\uparrow$ &$w$F$\uparrow$ &MAE $\downarrow$ \\
        \midrule[0.5pt]
        RGB&26.5M&83.1&0.818&0.054\\
        Depth&26.5M&43.8&0.715&0.061\\
        RGB+Depth& 26.7M &83.4&0.868& 0.048 \\
        \bottomrule[1pt]
    \end{tabular}
    \vspace{-8pt}
    \caption{\small Effect of different input modalities on capturing semantic categories and object shapes. Weighted F-measure ($w$F) and mean absolute error (MAE) are two common metrics for the foreground segmentation tasks~\cite{jiang2013salient,zhou2023specificity,yin2022camoformer}. }
    % \vspace{-5mm}
    \vspace{-5pt}
    \label{tab:class_seg}
\end{table}

\myPara{Inference Latency.}
Real-time inference speed is crucial for the practical deployment of RGB-D models across a wide range of downstream applications~\cite{chen2020sa_gate}.
%
% Thus, we test the inference latency of our \nameofmethod{} and other methods to evaluate their real-time potential.
Therefore, we evaluate the inference latency of \nameofmethod{} alongside other methods to assess their real-time potential.
To ensure a fair comparison, all tests are performed on the same hardware setup with a single 3090 RTX GPU, and the same image resolution of $480 \times 640$.
As shown in \tabref{tab:eff1}, \nameofmethod{} shows a good trade-off between speed and accuracy.

% \myPara{Generalization to Other Attention-base Model}

% To demonstrate the generalization capacity of our proposed geometry self-attention, we gradually transforming the attention mechanism of Swin Transformer to ours. 
% % 
% The results during transforming process are shown in.

\myPara{Discussion on the effect of RGB and depth.}
Semantic segmentation, assigning each pixel with a category label, can be seen as the combination of classification and segmentation of the objects.
Here, we explore how the two modalities contribute to capturing both semantic categories and object shapes, providing deeper insights into the design of our proposed geometry self-attention mechanism. 
To do so, we perform experiments on the LUSS~\cite{gao2022large} dataset, which provides segmentation annotations for 50K images from ImageNet~\cite{russakovsky2015imagenet}.
We divide the data into training, validation, and test sets, and train the model for both classification and foreground segmentation tasks. 
% 
% The experiment settings and adopted architectures are provided in the supplementary materials.
% 
As shown in \tabref{tab:class_seg}, we can see that the 3D geometry information within depth mainly helps the model segment the objects and slightly helps capture semantics.

\section{Conclusions}
We propose \nameofmethod{}, an RGBD vision backbone that incorporates an explicit geometry prior.
\nameofmethod{} leverages depth to model the geometry relationship between image patches and then uses this prior to allocate the weights of attention within the self-attention mechanism, called geometry self-attention.
Thanks to this tailored attention mechanism, our method achieves a more effective utilization of the depth modality. 
Experiments show that \nameofmethod{} produces better results than recent methods in RGB-D semantic segmentation with far less computational cost.

\myPara{Acknowledgment.}
This work was funded by NSFC (No. 62225604, 62176130), the Science and Technology Support Program of Tianjin, China (No. 23JCZDJC01050), and the Shenzhen Science and Technology Program (JCYJ20240813114237048). The Supercomputing Center of Nankai University partially supported computations.

{
    \small
    \bibliographystyle{ieeenat_fullname}
    \bibliography{main}
}

\end{document}

%% file: tabs/table_m2.tex
\begin{table*}[tp]
  \centering
% \vskip -1ex
    % \setlength\tabcolsep{6pt}
    % \vskip -1ex
    \setlength{\tabcolsep}{8.5pt}
    \footnotesize
    \vspace{-5pt}
    \centering
    \renewcommand{\arraystretch}{0.962}
    	\begin{tabular}{lcccccccc}
        \toprule
        \textbf{\multirow{2}{*}{Model}}  &\textbf{\multirow{2}{*}{Backbone}}   & \textbf{\multirow{2}{*}{Params}} &\multicolumn{3}{c}{NYUDepthv2} & \multicolumn{3}{c}{SUN-RGBD} \\ \cmidrule(lr){4-6}\cmidrule(lr){7-9}
         && &\textbf{Input size}& \textbf{Flops}& \textbf{mIoU}   &\textbf{Input size}& \textbf{Flops}& \textbf{mIoU} \\
        \midrule\midrule
        % ESANet$_{\rm 21}$~\cite{seichter2021efficient}&ResNet-34&31.2M&$480\times 640$&34.9G&50.3&$480\times 640$&34.9G&48.2\\ 
        TokenFusion$_{\rm 22}$~\cite{wang2022multimodal}& MiT-B2&26.0M&$480\times640$&55.2G&53.3&$530\times 730$&71.1G&50.3\\ 
        Omnivore$_{\rm 22}$~\cite{girdhar2022omnivore}&Swin-Tiny&29.1M&$480\times 640$&32.7G&49.7&$530\times730$&---&---\\ 
        DFormer$_{24}$~\cite{yin2023dformer} &DFormer-Tiny&6.0M&$480\times 640$& 11.7G&51.8&$530\times730$&15.0G&48.8\\
        DFormer$_{24}$~\cite{yin2023dformer} &DFormer-Small&18.7M&$480\times 640$ &25.6G&53.6 &$530\times730$&33.0G&50.0\\
        DFormer$_{24}$~\cite{yin2023dformer}&DFormer-Base&29.5M&$480\times 640$ &41.9G&55.6&$530\times730$&54.0G&51.2\\ 
        AsymFormer$_{\rm 24}$~\cite{du2024asymformer}                            & MiT-B0+ConvNeXt-Tiny                  & 33.0M                            & $480\times640$                 & 39.4G                        & 55.3                    & $530\times730$      &     52.6G           & 49.1                                   \\ 
         \rowcolor{gray!15}\highlight{$\bigstar$} \nameofmethod{}-S&\nameofmethod{}-Small&26.7M&$480\times640$&33.9G&\highlight{56.0}&$530\times730$&43.7G&\highlight{51.5}\\ \midrule
        SGNet$_{\rm 20}$~\cite{chen2021spatial_guided}&ResNet-101&64.7M&$480\times 640$ &108.5G& 51.1 &$530\times 730$&151.5G&48.6\\ 
        ShapeConv$_{\rm 21}$~\cite{cao2021shapeconv}&ResNext-101&86.8M&$480\times 640$ &124.6G& 51.3 &$530\times 730$&161.8G&48.6\\ 
        FRNet$_{\rm 22}$~\cite{zhou2022frnet}&ResNet-34&85.5M&$480\times 640$&115.6G&53.6&$530\times 730$&150.0G&51.8\\ 
         EMSANet$_{\rm 22}$~\cite{seichter2022efficient}&ResNet-34&46.9M&$480\times 640$&45.4G&51.0&$530\times 730$&58.6G&48.4\\ 
        % TransD-Fusion$_{\rm 22}$~\cite{wu2022transformer}&Swin-B&250.6M&$480\times 640$&---&55.5&$530\times 730$&---&51.9 &N/A\\
        
        TokenFusion$_{\rm 22}$~\cite{wang2022multimodal} &MiT-B3&45.9M&$480\times640$&94.4G&54.2&$530\times 730$&122.1G&51.4\\ 
        Omnivore$_{\rm 22}$~\cite{girdhar2022omnivore}&Swin-Small&51.3M&$480\times 640$&59.8G&52.7&$530\times730$&---&---\\ 
        CMX$_{\rm 22}$~\cite{zhang2022cmx}&MiT-B2&66.6M&$480\times 640$&67.6G&54.4&$530\times730$&86.3G&49.7\\ 
        DFormer$_{24}$~\cite{yin2023dformer}&DFormer-Large&39.0M&$480\times 640$ &65.7G&57.2&$530\times730$&84.5G&52.5\\ 
        GeminiFusion$_{\rm 24}$~\cite{jia2024geminifusion}                          & MiT-B3                             & 75.8M                            & $480\times640$                 &             138.2G                 & 56.8                           & $530\times730$      &       179.0G         & 52.7                                 \\ 
         \rowcolor{gray!15}\highlight{$\bigstar$} \nameofmethod{}-B&\nameofmethod{}-Base&53.9M&$480\times640$  &67.2G&\highlight{57.7}&$530\times730$&86.9G&\highlight{52.8}\\ \midrule
        %FCN~\cite{long2015fcn} & 29.2 & 49.2 \\
        % 3DGNN$_{\rm ICCV2017}$~\cite{qi20173d} &&&$480\times 640$&& 43.1 & - &&&45.9&-\\
        % Kong $et\ al.$~\cite{kong2018recurrent}&&&$480\times 640$ && 44.5 & 72.1 \\
        % LS-DeconvNet$_{2017}$~\cite{cheng2017locality}&&&$480\times 640$ && 45.9 & 71.9\\
        % CFN$_{\rm ICCV2017}$~\cite{CaRF17}&&&$480\times 640$ && 47.7 & -&&&48.1&- \\
        % ACNet$_{\rm 19}$~\cite{hu2019acnet}&ResNet-50&116.6M&$480\times 640$ &126.7G& 48.3 &$530\times 730$&163.9G&48.1 \\
        % PSD$_{\rm 20}$~\cite{zhou2020pattern}&ResNet-50&&$480\times 640$ &&51.0&&&50.6&N/A\\
        % CANet$_{\rm 20}$~\cite{zhou2020rgb}&ResNet-101&&$480\times 640$&&51.2&&&48.3&N/A\\
        % RDF$_{\rm ICCV2017}$~\cite{park2017rdfnet}&&ResNet-101&$480\times 640$ && 49.1 & 75.6&&&&\\
        
        % GLPNet$_{\rm 21}$~\cite{chen2021global}&ResNet-101&&&&54.6&&&51.2&N/A\\
        % FSFNet$_{\rm 21}$~\cite{su2021deep}&&&&&52.0&&&50.6&N/A\\
        % NANet$_{\rm 21}$~\cite{zhang2021non_aggregation}&&&$480\times 640$& & 52.3 &&&48.8&N/A \\
        SA-Gate$_{\rm 20}$~\cite{chen2020sa_gate}&ResNet-101&110.9M&$480\times 640$ &193.7G& 52.4 &$530\times 730$&250.1G&49.4 \\
        % DCANet$_{\rm 22}$~\cite{bai2022dcanet}&ResNet-101&&&&53.3&&&49.6&N/A\\
        % WTNet$_{\rm 22}$~\cite{fan2022rgb}&&&&&52.1&&&50.2&N/A\\
        % FAFNet$_{\rm 23}$~\cite{chen2023fafnet}&&&&&54.0&&&49.2&N/A\\
        % MGCNet$_{\rm 22}$~\cite{yang2022mgcnet}&&&&&54.5&&&51.5&N/A\\
        % RFNet$_{\rm 22}$~\cite{zhou2022rfnet}&&&&&53.5&&&50.7&N/A\\ \midrule
        % ZigZagNet$_{\rm 20}$~\cite{}
        CEN$_{\rm 20}$~\cite{wang2020deep}&ResNet-101&118.2M&$480\times 640$&618.7G&51.7&$530\times 730$&790.3G&50.2\\
        CEN$_{\rm 20}$~\cite{wang2020deep}&ResNet-152&133.9M&$480\times 640$&664.4G&52.5&$530\times 730$&849.7G&51.1\\
        
        PGDENet$_{\rm 22}$~\cite{zhou2022pgdenet}&ResNet-34&100.7M&$480\times 640$&178.8G&53.7&$530\times730$&229.1G&51.0\\
       
        MultiMAE$_{\rm 22}$~\cite{bachmann2022multimae}&ViT-Base&95.2M&$640\times640$&267.9G&56.0&$640\times 640$&267.9G&51.1$^{\dag}$\\

        Omnivore$_{\rm 22}$~\cite{girdhar2022omnivore}&Swin-Base&95.7M&$480\times 640$&109.3G&54.0&$530\times730$&---&---\\
        
        CMX$_{\rm 22}$~\cite{zhang2022cmx}&MiT-B4&139.9M&$480\times 640$&134.3G&56.3&$530\times730$&173.8G&52.1\\
        CMX$_{\rm 22}$~\cite{zhang2022cmx}&MiT-B5&181.1M&$480\times 640$&167.8G&56.9&$530\times730$&217.6G&52.4\\
        CMNext$_{\rm 23}$~\cite{zhang2023delivering}&MiT-B4&119.6M&$480\times640$&131.9G&56.9&$530\times730$&170.3G&51.9$^{\dag}$\\
        % \midrule
        % AMF$_{\rm 2022*}$&&ResNet-50&&&&&&49.6\\
        % PDCNet$_{\rm 2023*}$&&ResNet-101&&&&&&49.6\\
        % EMSAFormer$_{\rm 2023*}$&&&&&&&&48.8\\
        % EMSANet$_{\rm IJCNN2022*}$&&&&&53.3&&&48.5\\
        % LF$_{\rm 2023*}$&&Segformer-B2&&&&&&48.2\\
        % RedNet$_{\rm 2018*}$&&ResNet-50&&&&&&47.8\\
        % RDF-152$_{\rm ICCV2017*}$&&ResNet-152&&&50.1&&&47.7\\
        % CCL$_{\rm 2018*}$&&&&&&&&47.1\\
        % MMAF-Net-152$_{\rm 2019*}$&&&&&&&&47.0\\
        % UCTNet$_{\rm ECCV2022}$&&&&&57.6&&&&51.2&\\
        % CSNet$_{\rm JPRS2021}$&&&&&51.5&&&&52.8&\\
        % LWN$_{\rm TMM2021}$&&ResNet-152&&&51.5&&&&53.1&\\ %这个方法估计有问题

    GeminiFusion$_{\rm 24}$~\cite{jia2024geminifusion}                          & MiT-B5                             & 137.2M                           & $480\times640$                 &           256.1G                   & 57.7                           & $530\times730$      &    332.4G            & \highlight{53.3}                                \\
         \rowcolor{gray!15}\highlight{$\bigstar$} \nameofmethod{}-L&\nameofmethod{}-Large&95.5M&$480\times640$  &124.1G&\highlight{58.4}&$530\times730$&160.5G&\highlight{53.3}\\
        
        \bottomrule
        \end{tabular}
        \vspace{-5pt}
    \caption{\small Results on NYU Depth V2~\cite{silberman2012nyu_dataset} and SUN-RGBD~\cite{song2015sun_rgbd}.
    Some methods do not report the results or settings on the SUN-RGBD datasets, so we reproduce them with the same training configs. $^{\dag}$ indicates that we follow the results from~\cite{yin2023dformer}. All the backbones are pretrained on ImageNet-1K. We split the models to three sets, \ie small scale, base scale, and large scale. We can see that our method receives the best results on both datasets.}\label{tab:rgbd_sota}
    
    \hspace{\fill}
    \hspace{\fill}
    \vspace{-13pt}
\end{table*}